\documentclass[10pt,twocolumn,letterpaper]{article}

\usepackage[pagenumbers]{cvpr} 

\usepackage{bbding}

\usepackage{algorithm}
\usepackage{graphicx}
\usepackage{float}
\usepackage{caption}
\usepackage{subcaption}
\usepackage{algpseudocode}
\usepackage{multirow}

\usepackage[dvipsnames]{xcolor}

\definecolor{cvprblue}{rgb}{0.21,0.49,0.74}
\usepackage[pagebackref,breaklinks,colorlinks,citecolor=cvprblue]{hyperref}

\def\paperID{*****} 
\def\confName{CVPR}
\def\confYear{2024}

\title{An Open and Comprehensive Pipeline \\ for Unified Object Grounding and Detection}

\author{Xiangyu Zhao$^{1}$\ \  \ \ Yicheng Chen$^{1}$\ \  \ \ Shilin Xu$^{1}$\ \  \ \ Xiangtai Li$^{1}$ \\
Xinjiang Wang$^{2}$\ \  \ \ Yining Li$^{1}$\ \  \ \ Haian Huang$^{1,}$\footnotemark[2] \\
{ $^{1}$Shanghai AI Lab} \ \   { $^{2}$SenseTime Research} \\
{\tt\small \{zhaoxiangyu, chenyicheng, huanghaian\}@pjlab.org.cn} 
}

\begin{document}

\twocolumn[{
\maketitle
\begin{center}
    \vspace{-1em}
    \includegraphics[width=\textwidth]{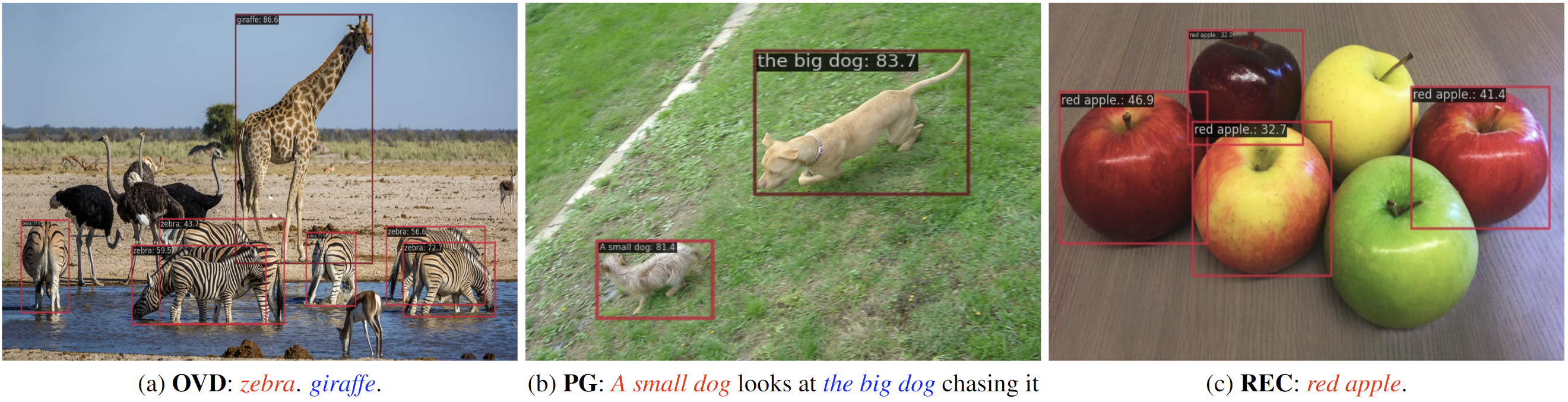}
    \captionof{figure}{(a)\textit{Open-Vocabulary Detection}(\textbf{OVD}). (b)\textit{Phrase Grounding}(\textbf{PG}). (c)\textit{Referring Expression Comprehension}(\textbf{REC}). }
\end{center}

}]

\begin{abstract}
Grounding-DINO is a state-of-the-art open-set detection model that tackles multiple vision tasks including Open-Vocabulary Detection (OVD), Phrase Grounding (PG), and Referring Expression Comprehension (REC).
Its effectiveness has led to its widespread adoption as a mainstream architecture for various downstream applications. 
However, despite its significance, the original Grounding-DINO model lacks comprehensive public technical details due to the unavailability of its training code.
To bridge this gap, we present \textbf{MM-Grounding-DINO}, an open-source, comprehensive, and user-friendly pipeline, which is built with the MMDetection toolbox.
It adopts abundant vision datasets for pre-training and various detection and grounding datasets for fine-tuning.  
We give a comprehensive analysis of each reported result and detailed settings for reproduction.
The extensive experiments on the benchmarks mentioned demonstrate that our MM-Grounding-DINO-Tiny outperforms the Grounding-DINO-Tiny baseline. 
We release all our models to the research community.
Codes and trained models are released at \url{https://github.com/open-mmlab/mmdetection/tree/main/configs/mm_grounding_dino}.

\end{abstract}
\footnotetext[2]{Project lead}
\label{sec:intro}
 \begin{figure}[ht]
    \begin{center}
        \includegraphics[width=\linewidth]{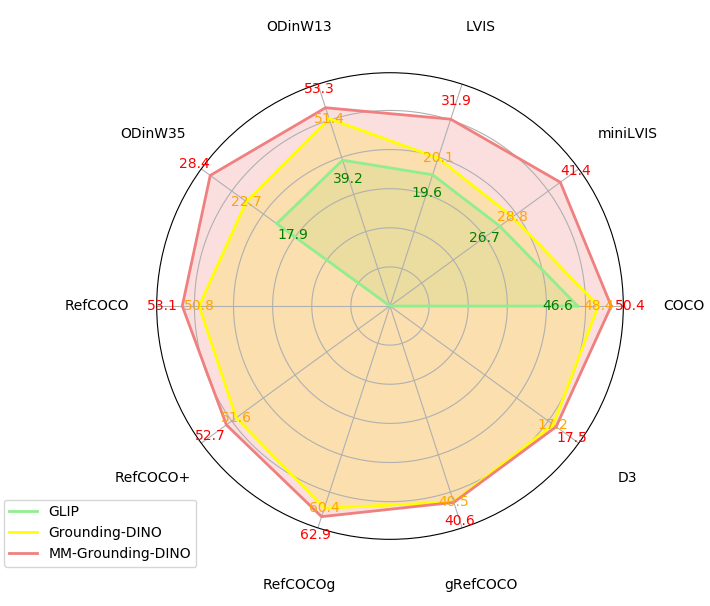}
    \end{center}
    \caption{Results on various benchmarks. MM-Grounding-DINO outperforms other grounding models on a broad range of tasks.}
    \label{fig:arch}
    \vspace{-1.5em}
\end{figure}
\section{Introduction}

The task of object detection typically involves inputting an image into a model to obtain proposals, which are then matched with text through multi-modal alignment, making it a key component of most state-of-the-art multi-modal understanding architectures. Presently, object detection can be subdivided into three sub-tasks according to the type of input text: Open-Vocabulary Detection (OVD), Phrase Grounding (PG), and Referring Expression Comprehension (REC). 

 Following zero-shot settings, OVD models are trained on base categories but require predicting both base and novel categories within a large-scale language vocabulary~\cite{survey1}. Phrase grounding task takes not only a category but a phrase that describes all the candidate categories as input and output corresponding boxes~\cite{radford2021learning}. The primary aim of REC task is to accurately identify the target designated by a given textual description and subsequently demarcate its position utilizing a bounding box~\cite{he2023grec}.

 In recent years, numerous vision grounding and detection models have been explored to solve the tasks above. Among these grounding models, Grounding-DINO~\cite{liu2023grounding} has been made as a main-stream architecture with superior performance. Based on a closed-set detector DINO~\cite{zhang2022dino}, Grounding-DINO-Large achieves state-of-the-art zero-shot performance on COCO~\cite{lin2015microsoft}(mAP 52.5) without any COCO training data. Grounding-DINO executes the integration of vision and language modality at various stages, encompassing feature enhancer, query selection module, and decoder. This profound fusion approach significantly enhances the detection of objects in an open-set context and DETR-based structure makes it an end-to-end network without any hard-crafted module.
 
Given that Grounding-DINO has demonstrated superior precision across the aforementioned three downstream tasks, yet is not entirely open-source (with only test and demo codes available), we rebuild the Grounding-DINO model utilizing the MMDetection toolbox~\cite{mmdetection} within the OpenMMLab project, adhering to the official test codes of Grounding-DINO. The structure of the model remains almost unchanged except for the modifications during initialization. Based on the Grounding-DINO framework, we propose to apply more datasets for pretraining, including COCO, Objects365~\cite{9009553}, GRIT~\cite{peng2023kosmos2}, V3Det~\cite{wang2023v3det}, RefCOCO~\cite{kazemzadeh-etal-2014-referitgame}, RefCOCO+~\cite{yu2016modeling}, RefCOCOg~\cite{mao2016generation}, GQA~\cite{hudson2018gqa} / Flickr30k Entities~\cite{flickrentitiesijcv}(combination also named as Golden-G dataset~\cite{kamath2021mdetr}), results in a stronger Grounding-DINO-based model which we call MM-Grounding-DINO. Since the Cap4M dataset~\cite{radford2021learning} used by Grounding-DINO is not open-source, we have opted for the GRIT and V3Det datasets as substitutes in our study.

We further extend all available benchmarks for OVD, PG and REC evaluation, including COCO, LVIS~\cite{gupta2019lvis}, RefCOCO/+/g, Flickr30k Entities, ODinW13/35~\cite{li2022elevater}, gRefCOCO~\cite{liu2023gres} and Description Detection Dataset($D^3$) ~\cite{xie2023described}. To our knowledge, we are the first to implement a framework that facilitates systematic evaluation across such an extensive array of datasets. All evaluation metrics are readily available in MMDetection. Pretrained with a large number of data, MM-Grounding-DINO-Tiny achieves zero-shot 50.6 mAP on COCO, 41.4 mAP on LVIS mini, and comprehensively surpasses Grounding-DINO-Tiny in REC task, detailed results are shown in Section \ref{sec:experiment}. We hope that our pipeline will serve as a valuable resource for further investigations in OVD, PG, and REC tasks.

The contributions of our paper are as follows:

\begin{enumerate}
    \item We propose MM-Grounding-DINO, a comprehensive and open-sourced grounding pipeline based on Grounding-DINO and pretrained with abundant vision datasets, which comprehensively address OVD, PG, and REC tasks.
    \item We take the lead in extending all available benchmarks for OVD, PG, and REC evaluation, including COCO, LVIS, RefCOCO/+/g, Flickr30K Entities, ODinW13/35, gRefCOCO and $D^3$. 
    All evaluation metrics are readily available in MMDetection.
    \item We extensively evaluate the transfer ability of our models by fine-tuning our model through a multitude of external special datasets.
    %
\end{enumerate}

\section{Approach}
\label{sec:baseline}
In this section, we introduce the model and datasets in detail. Unless otherwise specified, \textbf{MM-G} denotes MM-Grounding-DINO. \textbf{G-DINO} refers to Grounding-DINO. \textbf{O365} means Objects365 V1 and \textbf{GoldG} refers to the combination of GQA and Flickr30k Entities in the following sections.

\begin{figure*}[t]
    \begin{center}
        \includegraphics[width=\linewidth]{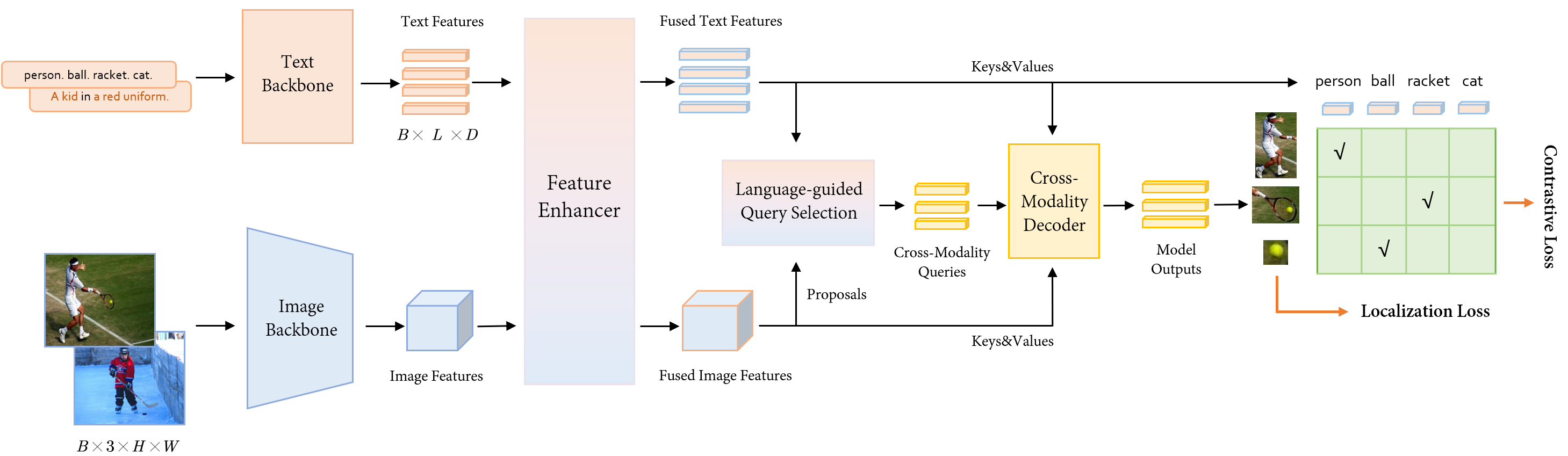}
    \end{center}
    \caption{Illustration of MM-Grounding-DINO. Given image and text description, a text backbone and an image backbone are first utilized to extract text and image features respectively. The images and text features are then fed into the feature enhancer module to perform deep cross-modality fusion. After fusing, a language-guided query selection module is employed to extract cross-modality queries from the image features. These cross-modality queries are subsequently inputted into a cross-modality decoder, which is designed to probe the desired features from the two modal features. The output queries generated by the final layer of the decoder are then utilized for the prediction of object boxes and corresponding phrases}
    \label{fig:arch}
\end{figure*}

\subsection{Model}

\begin{table*}[h]
    \caption{Variants of MM-G/G-DINO along with their corresponding pretraining datasets. The same sign(i.e., a, b, c) next to each variant denotes comparable groups. Since Cap4M is non-open-source, GRIT and V3Det are utilized as substitute datasets. VG: Visual Genome. RefC: RefCOCO/+/g. INB: ImageNetBoxes.}
    \label{tab:scale}
    \scalebox{0.9}{
    \begin{tabular}{l|c|c}
    \toprule
    Model & Training Datasets & Image Consumption \\
    \midrule
    MDETR & COCO, RefC, VG, GoldG & 52M (40 Ep $\times$ 1.3M Img) \\
    GLIP & O365, OpenImages, VG, INB, COCO, RefC, GoldG, Cap24M & 64M (64 Bs $\times$ 1M Iter) \\
    GLIPv2 (stage \uppercase\expandafter{\romannumeral2}) & COCO, LVIS, PhraseCut, Cap16M & 5.36M (24 Ep $\times$ 0.2M Img + 8 Ep $\times$ 0.07M Img) \\
    G-DINO-Tiny(a) & O365 & 18.3M (30 Ep $\times$ 128 Bs $\times$ 4755 Iter) \\
    G-DINO-Tiny(b) & O365, GoldG & 41.3M (30 Ep $\times$ 128 Bs $\times$ 10763 Iter) \\
    G-DINO-Tiny(c) & O365, GoldG, Cap4M & - \\
    MM-G-Tiny(a) & O365 & 18.3M (30 Ep $\times$ 128 Bs $\times$ 4755 Iter) \\
    MM-G-Tiny(b) & O365, GoldG & 41.3M (30 Ep $\times$ 128 Bs $\times$ 10763 Iter) \\
    MM-G-Tiny(c1) & O365, GoldG, GRIT & 56.3M (30 Ep $\times$ 128 Bs $\times$ 14669 Iter) \\
    MM-G-Tiny(c2) & O365, GoldG, V3Det & 46.8M (30 Ep $\times$ 128 Bs $\times$ 12196 Iter) \\
    MM-G-Tiny(c3) & O365, GoldG, GRIT, V3Det & 61.8M (30 Ep $\times$ 128Bs $\times$ 16102 Iter)\\
    MM-G-Large & COCO, RefC, O365V2, GoldG, GRIT, Open-Images, V3Det & - \\
    \bottomrule
    \end{tabular}
    }
\end{table*}

As we mentioned in Section~\ref{sec:intro}, our model is based on Grounding-DINO~\cite{liu2023grounding} and almost remains unchanged. Our framework is shown in Figure~\ref{fig:arch}. Given images with shape $[Batch size, 3, H, W]$ and text descriptions, our model can align the description with corresponding generated bounding boxes. The components of our model contain a text backbone for extracting text features, an image backbone for extracting image features, a feature enhancer for deeply fusing image and text features, a language-guided query selection module for query initialization, and a cross-modality decoder for box refinement. More details of the structure are drawn in ~\cite{liu2023grounding}. 

\noindent \textbf{Feature Extraction and Fusion}. Given an image-text pair, we employ an image backbone to extract image features at multiple scales, and concurrently, a text backbone is utilized for the extraction of text features. Then we fed both features into a feature enhancer module for cross-modality fusion. In the feature enhancer module, the text features and image features are first fused through a Bi-Attention Block containing both text-to-image cross-attention and image-to-text cross-attention layers. Then the fused text features and image features are additionally enhanced with vanilla self-attention and deformable self-attention layers followed by an FFN layer respectively, as drawn in Algorithm~\ref{Alg:fusion}. 


\begin{algorithm}
\caption{Feature Enhancer Layer}\label{Alg:fusion}
\begin{algorithmic}[1]
\State  fused\_image, fused\_text =  
\State  \indent BiAttentionBlock(image\_features, text\_features)
\State  fused\_text = 
\State  \indent FFN( SelfAttentionLayer(fused\_text) )
\State  fused\_image = 
\State  \indent FFN( DeformableAttentionLayer(fused\_image) )
\end{algorithmic}
\end{algorithm}

\noindent \textbf{Language-Guided Query Selection}. To optimize the utilization of text for guiding object detection, Grounding-DINO designed a language-guided query selection module. Language-guided query selection module selects $num\_query$ proposals based on the cosine similarity with input text features as decoder queries.
The parameter $num\_query$ denotes the number of queries fed into the decoder, and it has been configured to a value of 900 in our implementation, following DINO~\cite{zhang2022dino}.
The input queries for the decoder are composed of two components: the content part and the position part. 
The position part represents dynamic anchor boxes and is initialized based on the output of the language-guided query selection module, while the content part is initialized as an all-zero learnable query. 

\noindent \textbf{Cross-modality Decoder}. The cross-modality decoder layer in Grounding-DINO is designed to further incorporate text and image features for cross-modality learning. 
After self-attention, the architecture incorporates an image cross-attention layer, succeeded by a text cross-attention layer, and culminating in an FFN layer. Each decoder layer possesses an additional text cross-attention layer in comparison to the DINO decoder layer. 
This enhancement is necessitated by the requirement to inject textual information into queries, thereby facilitating the model's performance.


\noindent \textbf{Training Loss}. The L1 loss and the GIOU~\cite{rezatofighi2019generalized} loss are implemented for the box regression branch. Following GLIP~\cite{li2022grounded}, we utilize focal loss~\cite{lin2018focal} as a contrastive loss between the predicted boxes and language tokens for classification. Every predicted box would be multiplied with all language tokens to compute the similarity between them. Box regression and classification loss are jointly utilized for the computation of bipartite matching loss~\cite{carion2020endtoend}. Consistent with Grounding-DINO, we incorporate auxiliary loss for each decoder layer as well as the encoder outputs.”

\noindent \textbf{Difference}. The primary difference between MM-G and G-DINO lies in the \textit{contrastive embedding module}. 
%
Motivated by CLIP~\cite{radford2021learning}, we add bias while initializing the contrastive embedding module. 
This can significantly reduce the initial loss value and accelerate the convergence of our model. 
%

The implementation codes are shown in Algorithm \ref{Alg:contrastive}.
\begin{algorithm}
\caption{Contrastive Embedding}\label{Alg:contrastive}
\begin{algorithmic}[1]
\State res = visual\_feat @ text\_feat.transpose(-1, -2)
\State    \textcolor{blue}{res = res / math.sqrt(visual\_feat.shape[-1])}
\State    \textcolor{blue}{res = res + self.bias}

\end{algorithmic}
\end{algorithm}


\subsection{Datasets Preparation}
Our data format is motivated by the format in Open Grounding-DINO~\cite{Open_Grounding_Dino} and modified with the format in MMDetection. Since MM-Grounding-DINO is designed to address three tasks with datasets of different kinds of annotations, we divided the 15 datasets we used into three groups separately. Comprehensive details of the datasets are shown in Table \ref{tab:dataset}.  It's noteworthy that the entirety of data in GRIT, which exceeds 13 million, is not fully utilized per epoch during the training process. Instead, it is partitioned into segments of 500,000 for each epoch.

\noindent \textbf{OVD Datasets}. The datasets we use for training include \textit{COCO}~\cite{lin2015microsoft}, \textit{Objects365V1}~\cite{9009553}, \textit{Objects365V2}~\cite{9009553}, \textit{V3Det}~\cite{wang2023v3det}, \textit{Open-Images}, and the evaluation datasets contain \textit{COCO}, \textit{LVIS}~\cite{gupta2019lvis}, \textit{ODinW12/35}~\cite{li2022elevater}.

\noindent \textbf{PG Datasets}. The training datasets encompass \textit{GQA}~\cite{hudson2018gqa}, \textit{GRIT}~\cite{peng2023kosmos2}, \textit{Flickr30K Entities}~\cite{flickrentitiesijcv}, while\textit{ Flickr30K Entities} dataset is also used for evaluation.

\noindent \textbf{REC Datasets} The training datasets include \textit{RefCOCO}~\cite{kazemzadeh-etal-2014-referitgame}, \textit{RefCOCO+}~\cite{yu2016modeling}, \textit{RefCOCOg}~\cite{mao2016generation}. For evaluation, we utilize a broader set of datasets, which contain \textit{RefCOCO}, \textit{RefCOCO+}, \textit{RefCOCOg}, \textit{gRefCOCO}~\cite{liu2023gres}, and \textit{Description Detection Dataset($D^3$)}~\cite{xie2023described}.

\begin{table}[t]
    \begin{center}
    \caption{Comprehensive details of datasets utilized in MM-Grounding-DINO. Datasets denoted with an asterisk (*) can be utilized for both training and evaluation in our framework. Benchmarks are exclusively utilized for evaluation, while the remaining datasets are only for training.}
    \label{tab:dataset}
    \scalebox{0.7}{
    \begin{tabular}{l|cccccc}
    \toprule
    Dataset & Task & Images & Instances & categories  \\
    \midrule
    COCO\footnote[1]{Both utilized for training and evaluation}~\cite{lin2015microsoft} & OVD & 123K & 896K & 80 \\
    Objects365-V1~\cite{9009553} & OVD & 638K & 10M & 365\\
    Objects365-V2~\cite{9009553} & OVD & 1.7M & 25M & 365\\
    OpenImages-V6 & OVD & 1.5M & 14M & 600 \\
    V3Det~\cite{wang2023v3det} & OVD & 245K & 1753K & 13029\\
    Flickr30k Entities\footnote[1]~\cite{flickrentitiesijcv} & PG & 31K & 275K & - \\
    GQA~\cite{hudson2018gqa} & PG & 113K & - & - \\
    GRIT~\cite{peng2023kosmos2} & PG & 9M & 137M & - \\
    RefCOCO\footnote[1]~\cite{kazemzadeh-etal-2014-referitgame} & REC & 19K & 50K & - \\
    RefCOCO+\footnote[1]~\cite{yu2016modeling} & REC & 19K & 49K & - \\
    RefCOCOg\footnote[1]~\cite{mao2016generation} & REC & 26K & 54K & - \\
    \midrule
    LVIS~\cite{gupta2019lvis} & OVD Benchmark & 164K & 2M & 1000 \\
    ODinW~\cite{li2022elevater} & OVD Benchmark & 20K & 135K & 314 \\
    gRefCOCO~\cite{liu2023gres} & REC Benchmark & 19K & 60K & - \\
    $D^3$~\cite{xie2023described} & REC Benchmark & 10K & 18K & - \\
    \bottomrule
    \end{tabular}}
    \end{center}
\end{table}
\subsection{Training Settings}
\noindent \textbf{Input Rules of Text.} For OVD training, we concatenate all categories in detection datasets as a long string, like "\textit{People. Ball. Racket. Cat.}". For PG and REC tasks, following M-DETR~\cite{kamath2021mdetr}, during the pre-training phase, we annotate every object that is referred to within the text, which results in a slight modification in the model’s application for this task. For instance, during pre-training, given the caption "The woman wearing a blue dress standing next to the rose bush.", MM-Grounding-DINO would be trained to predict bounding boxes for all referred objects such as \textit{the woman}, \textit{the blue dress}, and \textit{the rose bush}. 

\noindent \textbf{Model Variants.} Similar to Grounding-DINO, we choose a well-pretrained BERT-based-uncased~\cite{devlin2019bert} model as our language encoder and Swin Transformer~\cite{liu2021swin} as image backbone. We compare different combinations of datasets in MM-G-tiny and G-DINO-Tiny. The selection of training datasets is contingent upon the scale of the image backbone, as shown in Table~\ref{tab:scale}.

\noindent \textbf{Data Augmentation}.
Besides random resize, random clip, and random flip, we also introduce random negative samples in data augmentation. We concatenate the categories or text descriptions, which are randomly sampled from other images as negative examples, with ground-truth descriptions serving as positive examples. This can effectively suppress the hallucination phenomena generated by the model, thus the model will not predict objects not exist in the image.

\noindent \textbf{Computing Resources}. We trained our MM-G-Tiny on 32 NVIDIA 3090 GPUs with a total batch size of 128 for 30 epochs. Since the computational cost of MM-G-Large is extremely high, MM-G-Large model is still in training.
\section{Main Results}
\label{sec:experiment}

\subsection{Zero-shot Transfer}
In zero-shot settings, MM-G models are initially trained on base datasets and subsequently assessed on novel datasets. Additionally, we present a set of results derived from fine-tuning to facilitate a comprehensive comparison of our model with Grounding-DINO. This approach ensures a robust evaluation of the model’s performance and its relative standing in the field.\\

\begin{table}[ht]
    \centering
    \caption{Results on COCO benchmark. All MM-G-T variants outperformed their counterparts, notably MM-G-T(c1) achieving 50.5 mAP in a zero-shot setting. }
    \label{tab:zero_coco}
    \begin{tabular}{lccc}
    \toprule
        Model & Backbone & Lr schd & COCO mAP \\ 
        \midrule
        GLIP & Swin-T & zero-shot & 46.6 \\ 
        G-DINO-T(a) & Swin-T & zero-shot & 46.7 \\
        G-DINO-T(b) & Swin-T & zero-shot & 48.1\\
        G-DINO-T(c) & Swin-T & zero-shot & 48.4 \\
        MM-G-T(a) & Swin-T & zero-shot & 48.5(+1.8) \\
        MM-G-T(b) & Swin-T & zero-shot & 50.4(+2.3) \\
        MM-G-T(c1) & Swin-T & zero-shot & 50.5(+2.1) \\
        MM-G-T(c2) & Swin-T & zero-shot & \textbf{50.6(+2.2)} \\
        MM-G-T(c3) & Swin-T & zero-shot & 50.4(+2.0)  \\ 
    \bottomrule
    \end{tabular}
\end{table}

\begin{table*}[ht]
    \centering
    \caption{Results on LVIS benchmark. All MM-G-T variants outperformed their counterparts. Significantly, MM-G-T(c3) reached 41.4 AP in the Mini Val zero-shot setting and showed a 17.3 improvement following 12 epochs of fine-tuning.}
    \label{tab: zero_LVIS}
    \resizebox{\textwidth}{!}{%
    \begin{tabular}{l|cccccccccc}
    \toprule
        Model & Backbone & Lr schd & \begin{tabular}[c]{@{}c@{}}MiniVal \\APr \end{tabular}& \begin{tabular}[c]{@{}c@{}}MiniVal \\APc \end{tabular}&\begin{tabular}[c]{@{}c@{}} MiniVal \\APf \end{tabular}&\begin{tabular}[c]{@{}c@{}} MiniVal \\AP \end{tabular}&\begin{tabular}[c]{@{}c@{}} Val1.0 \\APr \end{tabular}&\begin{tabular}[c]{@{}c@{}} Val1.0 \\APc \end{tabular}&\begin{tabular}[c]{@{}c@{}} Val1.0 \\APf \end{tabular}&\begin{tabular}[c]{@{}c@{}} Val1.0 \\AP \end{tabular} \\ \midrule
        GLIP & Swin-T & zero-shot & 18.1 & 21.2 & 33.1 & 26.7 & 10.8 & 14.7 & 29.0 & 19.6 \\
        G-DINO-T(c) & Swin-T & zero-shot & 18.8 & 24.2 & 34.7 & 28.8 & 10.1 & 15.3 & 29.9 & 20.1  \\ 
        MM-G-T(b) & Swin-T & zero-shot & 28.1 & 30.2 & 42.0 & 35.7 & 17.1 & 22.4 & 36.5 & 27.0  \\ 
        MM-G-T(c1) & Swin-T & zero-shot & 26.6 & 32.4 & 41.8 & \textbf{36.5(+7.7)} & 17.3 & 22.6 & 36.4 & \textbf{27.1(+7.0)}  \\ 
        MM-G-T(c2) & Swin-T & zero-shot & 33.0 & 35.6 & 45.9 & \textbf{40.5(+11.7)} & 17.3 & 22.6 & 36.4 & \textbf{27.1(+7.0)}  \\
        MM-G-T(c3) & Swin-T & zero-shot & \textbf{34.2} & \textbf{37.4} & \textbf{46.2} & \textbf{41.4(+12.6)} & \textbf{23.6} & \textbf{27.6} & \textbf{40.5} & \textbf{31.9(+11.8)}  \\
    \bottomrule
    \end{tabular}
    }%
\end{table*}
\noindent \textbf{COCO Benchmark}.  We conduct an evaluation of MM-Grounding-DINO pretrained on O365 dataset and other PG/REC datasets. Following Grounding-DINO, COCO dataset is utilized for establishing a zero-shot learning baseline. We compare MM-Grounding-DINO-Tiny with Grounding-DINO-Tiny in Table \ref{tab:zero_coco}. It's shown in the result that even MM-G(a) trained with O365 only(mAP 48.5) can outperform G-DINO(c) trained with O365, Gold-G, and Cap4M(mAP 48.4), which proves the efficiency of our model. Trained with objects365, Gold-G and GRIT,  MM-G-T(c) demonstrates a performance of \textbf{50.5 mAP}, which improves 2.1 AP over G-DINO(c) on COCO benchmark. This is achieved without the model being exposed to any COCO images during training, and GRIT data we use() is even less than Cap4M(4M). There are two potential explanations for this:
\begin{itemize}
    \item Our training strategy, particularly the additional bias during initialization, aids in the convergence of the model.
    \item  O365 dataset encompasses the categories of COCO dataset. Consequently, our model has been extensively trained on the O365 dataset and naturally exhibits improved accuracy on the COCO dataset. This assertion is indirectly validated by the comparatively lower performance observed when the model is evaluated on other datasets.
\end{itemize}
It has also been observed that the incorporation of V3Det dataset does not contribute positively to the COCO zero-shot evaluation, and may even have detrimental effects. \\

\noindent \textbf{LVIS Benchmark}. LVIS dataset constitutes a long-tail detection dataset, encompassing in more than 1000 distinct categories for evaluation. Following Grounding-DINO, LVIS is also utilized for zero-shot OVD evaluation. We compare MM-Grounding-DINO-Tiny with Grounding-DINO-Tiny in Table \ref{tab: zero_LVIS}. We observe that despite MM-G(a) being trained by O365 and GoldG in the absence of Cap4M, it still manages to surpass G-DINO(c) by +6.9AP on both LVIS MiniVal and Val. MM-G(c1) surpasses G-DINO(c) by +7.7AP on MiniVal and +7.0AP on Val, upon the addition of V3Det, MM-G(c3) experiences a substantial improvement of nearly 5 AP, reaching \textbf{41.4 mAP on MiniVal} and \textbf{31.9 mAP on Val}, which surpasses G-DINO(c) by a significant \textbf{+12.6 AP} on MiniVal and \textbf{+11.8 AP} on Val! The potential reasons could be categorized into two aspects:
\begin{itemize}
    \item Model exhibits more comprehensive training on LVIS categories vocabulary.
    \item V3Det includes more than 13k categories which may cover a large part of LVIS's categories, a similar conclusion is also drawn in \cite{yang2023recognize}.\\
\end{itemize}

\begin{table*}[!ht]
    \centering
    \caption{Results on RefCOCO/+/g. The training details of Grounding-DINO have not been released, so the learning schedule for fine-tuning is unknown. }
    \label{tab: zero_ref}
    \begin{tabular}{l|ccccccccccc}
    \toprule
        Method & Backbone & Setting & \multicolumn{3}{c}{RefCOCO} & \multicolumn{3}{c}{RefCOCO+} & \multicolumn{2}{c}{RefCOCOg} \\
        & & & val & testA & testB & val & testA & testB & val & test \\
        \hline
        G-DINO-T(c) & Swin-T & zero-shot & 50.8 & 57.4 & 45.0 & 51.6 & 57.3 & 46.4 & 60.4 & 59.7 \\
        MM-G-T(b) & Swin-T & zero-shot & 53.1 & \textbf{59.7} & 46.4 & 53.1 & 58.9 & 47.9 & 61.2 & 61.1 \\
        MM-G-T(c1) & Swin-T & zero-shot & \textbf{53.4} & 58.8 & \textbf{46.8} & \textbf{53.5} & \textbf{59.0} & 47.9 & 62.7 & 62.6 \\
        MM-G-T(c2) & Swin-T & zero-shot & 52.1 & 58.4 & 45.4 & 52.5 & 58.2 & 46.9 & 61.7 & 61.0 \\
        MM-G-T(c3) & Swin-T & zero-shot & 53.1 & 59.1 & \textbf{46.8} & 52.7 & 58.7 & \textbf{48.4} & \textbf{62.9} & \textbf{62.9} \\
        \midrule
        G-DINO-T(c) & Swin-T & - & 89.2 & \textbf{91.9} & 86.0 & 81.1 & 87.4 & \textbf{74.7} & 84.2 & 84.9 \\
        MM-G-T(c3)  & Swin-T & 5e & \textbf{89.5} & 91.4 & \textbf{86.6} & \textbf{82.1} & \textbf{87.5} & 74.0 & \textbf{85.5} & \textbf{85.8} \\
    \bottomrule
    \end{tabular}
\end{table*}

\noindent \textbf{ODinW Benchmark}. ODinW (Object Detection in the Wild) benchmark represents a more rigorous benchmark designed to assess model performance within real-world contexts. It consists of 35 object detection datasets, each of which is augmented with external knowledge. We utilize ODinW13/35 to evaluate the transferability of our model, summary results are shown in Table \ref{tab: zero_ODin}. Our MM-G-T(c3)  demonstrates superior performance over G-DINO-T(c) and achieves scores of \textbf{53.3 mAP} and \textbf{28.4 mAP} on ODinW13 and ODinW35 respectively, which proves the robust transferability of our model. It is evident that a wide vocabulary holds substantial significance for ODinW datasets. Upon the integration of V3Det, the model’s performance experienced a substantial enhancement. The primary reason for this improvement is that V3Det encompasses a broader range of categories within ODinW. Detailed results of each sub-dataset are shown in Appendix \ref{app:Odin}.\\
\begin{table}[h!]
    \centering
    \caption{Zero-shot domain transfer on ODinW. }
    \label{tab: zero_ODin}
    \begin{tabular}{l|ccc}
    \toprule
        Model & Backbone & ODinW13 & ODinW35 \\ \midrule
        G-DINO-T(c) &Swin-T & 51.4 & 22.7 \\ 
        MM-G-T(b) &Swin-T& 45.3 & 20.2 \\ 
        MM-G-T(c1) &Swin-T& 51.1 & 22.8 \\ 
        MM-G-T(c2) &Swin-T& 51.1 & 22.8 \\
        MM-G-T(c3) &Swin-T& \textbf{53.3(+1.9)} & \textbf{28.4(+5.7)} \\ 
        \bottomrule
    \end{tabular}
\end{table}

\begin{table}[!ht]
    \centering
    \caption{Results on gRefCOCO benchmark. Note that the threshold value is set to 0.6. More results can refer to Appendix \ref{app:gref}. }
    \label{tab: zero_gref}
    \resizebox{23.5em}{3.4em}{%
    \begin{tabular}{l|ccccc}
    \toprule
        \multirow{3}{*}{Method} & \multirow{3}{*}{Backbone} & \multicolumn{3}{c}{gRefCOCO} \\
        &&\multicolumn{3}{c}{Pr\@(F1=1, IoU$\geqslant$0.5) / N-acc}\\
        & & val & testA & testB \\
        \midrule
        G-DINO-T(c) & Swin-T & 40.5/\textbf{83.8} & \textbf{29.3}/82.9 & 30.0/86.1 \\
        MM-G-T(c3) & Swin-T & \textbf{40.6}/83.1 & 29.2/\textbf{84.3} & \textbf{31.6}/\textbf{96.7} \\
    \bottomrule
    \end{tabular}
    }%
\end{table}

\noindent \textbf{RefCOCO/+/g and gRefCOCO Benchmark}. We also evaluate MM-G's zero-shot ability on REC task. RefCOCO, RefCOCO+, and RefCOCOg are established for REC evaluation, results are shown in Table \ref{tab: zero_ref}. Compared with RefCOCO, gRefCOCO broadens its scope to encompass multi-target expressions, which entail the specification of multiple target objects via a single expression. Additionally, gRefCOCO accommodates no-target expressions that do not refer to any object within the image. This augmentation introduces a markedly elevated degree of versatility to input expressions, consequently enhancing the practicality and robustness of REC in real-world applications. We also conduct an evaluation on gRefCOCO benchmark to assess the zero-shot capabilities of REC, with the results being presented in Table \ref{tab: zero_gref}. Our model is able to surpass the baseline across all zero-shot evaluation metrics on RefCOCO, and can either surpass or approximately equal G-DINO on gRefCOCO. From the results, it can be inferred that V3Det dataset can not provide any benefit for REC task. \\


\noindent \textbf{Description Detection Dataset($\mathbf{D^3}$) Benchmark}. $D^3$ is characterized by its flexible language expressions, ranging from concise category names to extensive descriptions, and it ensures comprehensive annotation of all objects described across all images without omission. Sentences in $D^3$ are slightly longer than an ordinary word, therefore, it does not require a high level of understanding ability from the model. In fact, it leans more towards OVD task. In addition, there are 24,282 positive object-text pairs and 7,788,626 negative pairs in $D^3$, which imposes a stringent demand on the model’s ability to distinguish negative objects. We report our results in Table \ref{tab:DOD}. From the results, we observe that MM-G-T(c1) with GRIT and G-DINO-T(c) trained with Cap4M have demonstrated comparable performance. In particular, MM-G-T(c1) exhibits advancements in long sentences, while G-DINO-T(c) shows progress when dealing with short sentences. This will be elaborated in detail in Section \ref{sec:GRIT}. After incorporating V3Det, which contains a large number of precise annotations, the performance of MM-G-T(c3) on short sentences surpassed G-DINO-T(c) while the performance on long sentences gets worse. This is primarily due to the fact that the majority of text annotations in V3Det are short sentences.

\begin{table*}[ht]
    \centering
    \caption{Zero-shot transfer on $D^3$. FULL, PRES, and ABS denote \textit{evaluation on all descriptions}, \textit{presence descriptions only}, and \textit{absence descriptions only} respectively. s/m/l/vl denote \textit{short}, \textit{middle}, \textit{long} and \textit{very long}.}
    \label{tab:DOD}
    \resizebox{\textwidth}{!}{%
    \begin{tabular}{l|lccccc}
    \toprule
        Method & mode & G-DINO-T(c) & MM-G-T(b) & MM-G-T(c1) & MM-G-T(c2) & MM-G-T(c3) \\ 
        \midrule
        \multirow{2}{*}{FULL/s/m/l/vl} & concat & 17.2/18.0/\textbf{18.7}/14.8/\textbf{16.3} & 15.6/17.3/16.7/14.3/13.1 & 17.0/17.7/18.0/\textbf{15.7}/15.7 &16.2/17.4/16.8/14.9/15.4 & \textbf{17.5}/\textbf{23.4}/18.3/14.7/13.8 \\ 
         & parallel & 22.3/\textbf{28.2}/24.8/19.1/13.9 & 21.7/24.7/24.0/20.2/13.7 & 22.5/25.6/25.1/\textbf{20.5}/\textbf{14.9} &22.3/25.6/24.5/20.6/14.7 & \textbf{22.9}/28.1/\textbf{25.4}/20.4/14.4 \\ \hline
        \multirow{2}{*}{PRES/s/m/l/vl} & concat & 17.8/18.3/\textbf{19.2}/15.2/17.3 & 16.4/18.4/17.3/14.5/14.2 & 17.9/19.0/18.3/\textbf{16.5}/\textbf{17.5} &16.6/18.8/17.1/15.1/15.0 & \textbf{18.0}/\textbf{23.7}/18.6/15.4/13.3 \\ 
         & parallel & 21.0/\textbf{27.0}/22.8/17.5/12.5 & 21.3/25.5/22.8/\textbf{19.2}/12.9 & 21.5/25.2/23.0/19.0/\textbf{15.0} &21.6/25.7/23.0/19.5/14.8 & \textbf{21.9}/27.4/\textbf{23.2}/19.1/14.2 \\ \hline
        \multirow{2}{*}{ABS/s/m/l/vl} & concat & 15.4/17.1/16.4/\textbf{13.6}/\textbf{14.9} & 13.4/13.4/14.5/13.5/11.9 & 14.5/13.1/16.7/\textbf{13.6}/13.3 &14.8/12.5/15.6/14.3/15.8 & \textbf{15.9}/\textbf{22.2}/\textbf{17.1}/12.5/14.4 \\ 
         & parallel & \textbf{26.0}/\textbf{32.0}/33.0/23.6/\textbf{15.5} & 22.8/22.2/28.7/22.9/14.7 & 25.6/26.8/33.9/\textbf{24.5}/14.7 &24.1/24.9/30.7/23.8/14.7 & \textbf{26.0}/30.3/\textbf{34.1}/23.9/14.6 \\ 
    \bottomrule
    \end{tabular}
    }%
\end{table*}

\begin{table*}[ht]
    \centering
    \caption{Fine-tune Results on LVIS benchmark. }
    \label{tab:finetune_LVIS}
    \resizebox{\textwidth}{!}{%
    \begin{tabular}{l|cccccccccc}
    \toprule
        Model & Backbone & Setting & \begin{tabular}[c]{@{}c@{}}MiniVal \\APr \end{tabular}& \begin{tabular}[c]{@{}c@{}}MiniVal \\APc \end{tabular}&\begin{tabular}[c]{@{}c@{}} MiniVal \\APf \end{tabular}&\begin{tabular}[c]{@{}c@{}} MiniVal \\AP \end{tabular}&\begin{tabular}[c]{@{}c@{}} Val1.0 \\APr \end{tabular}&\begin{tabular}[c]{@{}c@{}} Val1.0 \\APc \end{tabular}&\begin{tabular}[c]{@{}c@{}} Val1.0 \\APf \end{tabular}&\begin{tabular}[c]{@{}c@{}} Val1.0 \\AP \end{tabular} \\ \midrule
        MM-G-T(c3) & Swin-T & zero-shot & 34.2 & 37.4 & 46.2 & 41.4 & 23.6 & 27.6 & 40.5 & 31.9  \\
        MM-G-T(c3) & Swin-T & open-set 1x & \textbf{50.7(+16.5)} & \textbf{58.8(+21.4)} & \textbf{60.1(+13.9)} & \textbf{58.7(+17.3)} & \textbf{45.2(+21.6)} & \textbf{50.2(+12.6)} & \textbf{56.1(+15.6)} & \textbf{51.7(+19.8)} \\
        MM-G-T(c3) & Swin-T & open vocabulary 1x & \textbf{43.2(+9.0)} & \textbf{57.4(+20.0)} & \textbf{59.3(+13.1)} & \textbf{57.1(+15.7)} & - & - & - & - \\
    \bottomrule
    \end{tabular}
    }%
\end{table*}

\begin{table}[ht]
    \centering
    \caption{Fine-tune Results on COCO benchmark. Both close-set and open-set fine-tuning achieve a large improvement over the pre-trained model. }
    \label{tab:finetune_coco}
    \begin{tabular}{lcccc}
    \toprule
        Model & Backbone & Setting & mAP \\ 
        \midrule
        GLIP & Swin-T & zero-shot & 46.6 \\
        G-DINO-T(c) & Swin-T & zero-shot & 48.4 \\
        
        MM-G-T(c1) & Swin-T & zero-shot & 50.5(+2.1) \\
        MM-G-T(c2) & Swin-T & zero-shot & \textbf{50.6(+2.2)} \\
        MM-G-T(c3) & Swin-T & zero-shot & 50.4(+2.0)  \\ 
        \midrule
        Faster R-CNN & R-50 & close-set 1x & 37.4 \\
        Cascade R-CNN & R-50 & close-set 1x & 40.3 \\
        ATSS & R-50 & close-set 1x & 39.4 \\
        TOOD & R-50 & close-set 1x & 42.4 \\
        DINO & R-50 & close-set 1x & 50.1 \\
        GLIP & Swin-T & close-set 1x & 55.4 \\
        G-DINO-T(c) & Swin-T & close-set 1x & 58.1 \\
        MM-G-T(c3) & Swin-T & close-set 1x & \textbf{58.2(+7.8)} \\
        \midrule
        MM-G-T(c3) & Swin-T & open-set 1x & \textbf{54.7(+4.3)} \\
        
    \bottomrule
    \end{tabular}
\end{table}

\subsection{Analysis for GRIT}
\label{sec:GRIT}
GRIT\cite{peng2023kosmos2} is a large dataset employed as our substitute for Cap4M created in GLIP~\cite{li2022grounded}, given that the latter is not open-source. However, as shown in the results above, the performance of GRIT doesn't meet our expectations. For OVD task, MM-G-T(c1) with GRIT only improve +0.1 AP on COCO in Table \ref{tab:zero_coco} and +0.1 AP(Val) on LVIS in Table \ref{tab: zero_LVIS} than MM-G-T(b) without GRIT. For REC task, the gain brought by GRIT is relatively low on RefCOCO and gRefCOCO in Table \ref{tab: zero_ref} and \ref{tab: zero_gref}. From our observation of images and annotations in GRIT, the primary reasons can be enumerated as follows:
\begin{itemize}
    \item The text annotation of GRIT comes from phrases or sentences extracted by spaCy\cite{honnibal2020spacy} from captions in COYO-700M and LAION-2B, including a large number of abstract phrases like human names, events, facilities, and Geo-Political entities, which could potentially lead to the misdirection of the model.
    \item In GRIT dataset, the majority of images are accompanied by a singular annotation. The single annotation encompasses a long sentence which is actually the whole caption of the image and a noisy box which approximately spans the full extent of the image.
\end{itemize}
However, it is noteworthy that the large-scale data of GRIT still serves a purpose. MM-G-T(c1) with GRIT surpass MM-G-T(b) by 5.8/2.6 AP on ODinW13/35 in Table \ref{tab: zero_ODin}, which is on par with G-DINO-T(c) pretrained with Cap4M. We thus  Observed from Table \ref{tab:DOD}, MM-G-T(c1) with GRIT and G-DINO-T(c) with Cap4M have demonstrated comparable performance on $D^3$. Fortuitously, the single long text annotation of GRIT contributes to the enhancement of MM-G-T(c1)'s performance on long sentences.

\begin{table*}[ht]
    \centering
    \caption{Fine-tune results on downstream tasks. * denotes 12 epochs of fine-tuning setting. \textdagger denotes 50 epochs of fine-tune setting. Under the same fine-tuning setting, our MM-Grounding-DINO outperforms previous methods by a large margin for most downstream tasks. The evaluation metric is box AP.}
    \label{tab:finetune}
    \begin{tabular}{l|l|c|ccccc}
    \toprule
        Method & Backbone & Fine-tune & RTTS\footnotemark[1] & RUOD\footnotemark[1] & Brain Tumor\footnotemark[2] & Cityscapes\footnotemark[2] & People in Painting\footnotemark[2] \\
    \midrule
        Faster R-CNN & R-50 & \Checkmark & 48.1 & 52.4 & 43.5 & 30.1 & 17.0 \\
        Cascade R-CNN & R-50 & \Checkmark & 50.8 & 55.3 & 46.2 & 31.8 & 18.0 \\
        ATSS & R-50 & \Checkmark & 48.2 & 55.7 & - & - & - \\
        TOOD & R-50 & \Checkmark & 50.8 & 57.4 & - & - & - \\
        DINO & R-50 & \Checkmark & - & - & 46.4 & 34.5 & 12.0 \\
        Cascade-DINO & R-50 & \Checkmark & - & - & \textbf{48.6} & 34.8 & 13.4 \\
        MM-GDINO & Swin-T & \XSolidBrush & 49.8 & 29.8 & 0.4 & 34.2 & 23.1 \\
        MM-GDINO & Swin-T & \Checkmark & \textbf{69.1} & \textbf{65.5} & 47.5 & \textbf{51.5} & \textbf{38.9} \\
    \bottomrule
    \end{tabular}
\end{table*}

\subsection{Validation through Fine-tuning}

The default fine-tuning in this report is based on MM-G-T(c3) pre-trained model.

\subsubsection{Fine-tuning on COCO/LVIS}
\label{sec:finetune_coco}

\noindent \textbf{Fine-tune on COCO. } We implemented three mainstream fine-tuning approaches with MM-Grounding-DINO to thoroughly evaluate its capabilities: close-set fine-tuning, open-set continuing pretraining fine-tuning, and open-vocabulary fine-tuning. The latter two fine-tuning methods are designed to keep model's generalizability while enhancing performance on the COCO dataset.
\begin{itemize}
    \item In close-set fine-tuning, we fine-tuned our model using a close-set algorithm, optimizing specifically for COCO dataset. Post fine-tuning, the textual input was restricted to COCO categories. 
    \item In the open-set continuing pretraining fine-tuning, we derived two distinct methods based on the same training strategy in pretraining phase. The first involves lowering the learning rate and freezing certain modules, then continue training on COCO dataset. The second method combines the COCO dataset with other pre-training datasets of MM-G-T(c3) for continue training. 
    \item For open-vocabulary fine-tuning, we categorized the dataset into base and novel categories. During fine-tuning, only base categories were utilized. Subsequently, we evaluated the model's performance across both base and novel categories.
\end{itemize}

As shown in Table \ref{tab:finetune_coco}, MM-G-T significantly improved performance on the COCO dataset through both close-set fine-tuning and open-set continuing pretraining fine-tuning. Notably, MM-G-T achieved a 7.8 mAP increase after 12 epochs of close-set fine-tuning, reaching 58.2 mAP. For more results regarding open-vocabulary fine-tuning, please refer to Table \ref{tab:ovd_finetune_coco} in Appendix \ref{app:coco}.\\

\noindent \textbf{Fine-tune on LVIS. } LVIS dataset, characterized by its long-tail distribution, encompasses 1203 categories. Given this extensive categorization, we exclusively employed open-set continue pretraining fine-tuning and open vocabulary fine-tuning for this dataset.

As illustrated in Table \ref{tab:finetune_LVIS}, the open-set continuing pretraining fine-tuning significantly enhanced MM-G-T's performance. Notably, MM-G-T achieved a substantial increase of 9.0 mAP in the Apr metric for Mini LVIS.

\subsubsection{Fine-tuning on REC}
\noindent \textbf{Fine-tune for RefCOCO/+/g. } We further evaluate our model by fine-tuning on REC task as detailed in Table \ref{tab: zero_ref}. Following MDETR~\cite{kamath2021mdetr}, we adapted the fine-tuning phase to phrase grounding, consistent with the pre-training. The results, presented in Table \ref{tab: zero_ref}, indicate a marked improvement in REC task performance after just 5 epochs of fine-tuning. This suggests that the current RefCOCO/+/g dataset, along with its evaluation metric, might be overly simplistic. Utilizing phrase grounding for fine-tuning on this task still leads to significant enhancements. We look forward to the emergence of a more robust and rigorous evaluation metric for further advancing REC task proficiency.

\subsubsection{Fine-tuning on Downstream Tasks}
To comprehensively show the generalizability of MM-Grounding-DINO, we extended its evaluation to various downstream tasks. In the fine-tuning settings, models are initially trained on expansive datasets and are then specifically trained using the training sets from the respective downstream tasks.

\noindent \textbf{Object Detection in the Haze. } Our study utilized the Real-world Task-driven Testing Set (RTTS), comprising 4,322 real-world hazy images predominantly featuring traffic and driving scenarios\cite{li2019benchmarking}. The RTTS dataset encompasses a variety of common categories within hazy conditions, offering an apt platform to access our model's efficacy and generalizability in diverse environments. We adopted the same dehazing and detection joint pipeline proposed in the benchmark. Impressively, MM-Grounding-DINO reached 69.1 AP after 12 epochs of fine-tuning, surpassing previous standards by a large margin as shown in Table \ref{tab:finetune}.

\noindent \textbf{Object Detection Underwater. } In this study, we evaluated the performance of MM-Grounding-DINO on the Real-world Underwater Object Detection dataset (RUOD)\cite{Fu2023reth}. This dataset comprises 14,000 high-resolution images with 74,903 labeled instances. Characterized by its diverse range of categories, object scales, image scales, object densities, and category densities, the dataset also introduces a range of underwater challenges. These include haze-like effects, color casts, light interference, and complex marine objects. This evaluation leveraged the RUOD dataset to ascertain the capabilities of our model in a distinct image domain, while simultaneously engaging with a subset of commonly encountered objects.

Table \ref{tab:finetune} shows that in the zero-shot setting, the MM-Grounding-DINO achieved a mAP of 29.8, primarily due to distribution mismatches between the training dataset, which mainly consists of terrestrial images, and the RUOD. However, after 12 epochs of fine-tuning, the model showed an improvement of 35.7 mAP, thereby setting a new benchmark. This performance surpasses the previous state-of-the-art by 8.1 mAP.

\noindent \textbf{Object Detection for Brain Tumor. } We further extended our evaluation to the medical domain, utilizing the Brain tumor dataset\cite{aa2022brain}. Notably, this dataset is unique in its labeling approach, as it only utilizes numerical identifiers without providing descriptive label information. As detailed in Table \ref{tab:finetune}, the performance of MM-Grounding-DINO underperformed Cascade-DINO\cite{ye2023cascadedetr}. We hypothesize that the relatively suboptimal results of our model could be attributed to the challenges posed by the dataset's reliance on purely numerical labels, especially in scenarios where the textual context is completely unknown.

\noindent \textbf{Object Detection for Cityscapes. } Cityscapes\cite{cordts2016cityscapes} is an extensive urban street scene collection, comprising 3k training images and 500 validation images. It features a broad and varied set of stereo video sequences captured in the streets of 50 different cities, accompanied by high-quality, pixel-level annotations. This dataset assessed our model's performance in recognizing common objects encountered in daily life. Notably, in Table \ref{tab:finetune}, we can observe that our pretrained MM-Grounding-DINO already performed on par with fine-tuned models without the need for any dataset-specific training. After 50 epochs of fine-tuning, it had an improvement of 17.3 mAP, reaching new state-of-the-art.

\noindent \textbf{Object Detection for People in Painting. } People in Paintings \cite{people-in-paintings_dataset} was originally created by Raya AI as a part of RF100, an initiative to establish a new object detection benchmark for model generalizability. The annotations in this dataset exclusively pertain to the figures depicted in paintings. As illustrated in Table \ref{tab:finetune}, our MM-Grounding-DINO model has already outperformed the performance of fine-tuned models in a zero-shot setting. Following a fine-tuning of 50 epochs, it demonstrated a significant improvement, achieving an increase of +15.8AP, setting a new benchmark of 38.9 mAP. 
\section{Conclusion}
\label{sec:conc}
In this paper, We propose MM-Grounding-DINO, a comprehensive and open-sourced grounding baseline based on Grounding-DINO and pretrained with abundant vision datasets, and comprehensively address OVD, PG, and REC tasks. We extend all available benchmarks for OVD, PG, and REC evaluation, and all evaluation metrics are readily available in MMDetection. The extensive experiments on the benchmarks mentioned demonstrate that our MM-Grounding-DINO outperforms (or is on par with) the Grounding-DINO baseline. We hope that our pipeline will serve as a valuable resource for further investigations in grounding and detection tasks.

{
    \small
    \bibliographystyle{ieeenat_fullname}
    \bibliography{main}
}
\appendix
\clearpage
\setcounter{page}{1}
\maketitlesupplementary

\section{More Results}
\label{sec:More_Results}

\subsection{Detailed Results on gRefCOCO}
\label{app:gref}
In our experiments, we initially set the default threshold to 0.7 following ~\cite{liu2023gres}. Then we conducted an extensive series of tests with varying threshold values. The impact of these different thresholds on our results is detailed in Table\ref{tab:more_gref}. We observed distinct effects of the threshold adjustments on the outputs. Specifically, a threshold of 0.8 yielded the highest F1 score for the validation set. In contrast, for both test sets A and B, a lower threshold of 0.5 proved more effective. This leads to an anticipation for the development of a more robust evaluation metric for this dataset. It's noteworthy that following the fine-tuning process (threshold is set to 0.7), all subsets of gRefCOCO demonstrated significant improvements.

\subsection{Detailed Results on Flickr30K Entities}
As shown in Table\ref{tab:flick}, MM-G-T(c) exhibits lower performance compared to G-DINO-T on Flickr30K Entities. Given that the GoldG dataset includes images from Flickr30K Entities, it's important to note that these results do not represent a zero-shot scenario. The observed performance differences could be attributed to variations in training strategies and settings.

\subsection{Detailed Results on ODinW datasets}
\label{app:Odin}
We provide the details of the 35 datasets we use in Table\ref{tab:ODinW35}. Considering the rarity of categories within the ODinW13/35 dataset, the additional concepts brought by GRIT and V3Det datasets prove to be beneficial. 

\subsection{Open-Vocabulary Fine-tuning on COCO}
\label{app:coco}

As elaborated in section\ref{sec:finetune_coco}, the results in Table\ref{tab:ovd_finetune_coco} show that despite fine-tuning solely on base categories, there is an observable +1.5 mAP enhancement in the novel categories. This finding demonstrates the effectiveness of open-vocabulary fine-tuning in preserving the model's generalizability.

\begin{table}[ht]
    \centering
    \caption{Results on gRefCOCO benchmark with different threshold values.}
    \label{tab:more_gref}
    \resizebox{22em}{!}{
    \begin{tabular}{l|ccccc}
    \toprule
        \multirow{3}{*}{Method} & \multirow{3}{*}{Threshold} & \multirow{3}{*}{Setting} & \multicolumn{3}{c}{gRefCOCO} \\
        &&&\multicolumn{3}{c}{Pr\@(F1=1, IoU$\geqslant$0.5)}\\
        & & &val & testA & testB \\
        \midrule
        G-DINO-T(c) & 0.5 & zero-shot & 39.3 & 31.9 & 30.0 \\
        G-DINO-T(c) & 0.6 & zero-shot & 40.5 & 29.3 & 30.0 \\
        G-DINO-T(c) & 0.7 & zero-shot & 41.3 & 27.2 & 29.7 \\
        G-DINO-T(c) & 0.8 & zero-shot & 41.5 & 25.1 & 29.1 \\
        MM-G-T(c3) & 0.5 & zero-shot & 39.4 & 33.1 & 33.0 \\
        MM-G-T(c3) & 0.6 & zero-shot & 40.6 & 29.2 & 31.6 \\
        MM-G-T(c3) & 0.7 & zero-shot & 41.0 & 26.1 & 30.4 \\
        MM-G-T(c3) & 0.8 & zero-shot & 41.1 & 23.8 & 29.5 \\
        \midrule
        MM-G-T(c3) & 0.7 & fine-tune & \textbf{45.1(+4.1)} & \textbf{42.5(+16.4)} & \textbf{40.3(+9.9)} \\
    \bottomrule
    \end{tabular}
    }
\end{table}

\begin{figure}[t]
    \begin{center}
        \includegraphics[width=\linewidth]{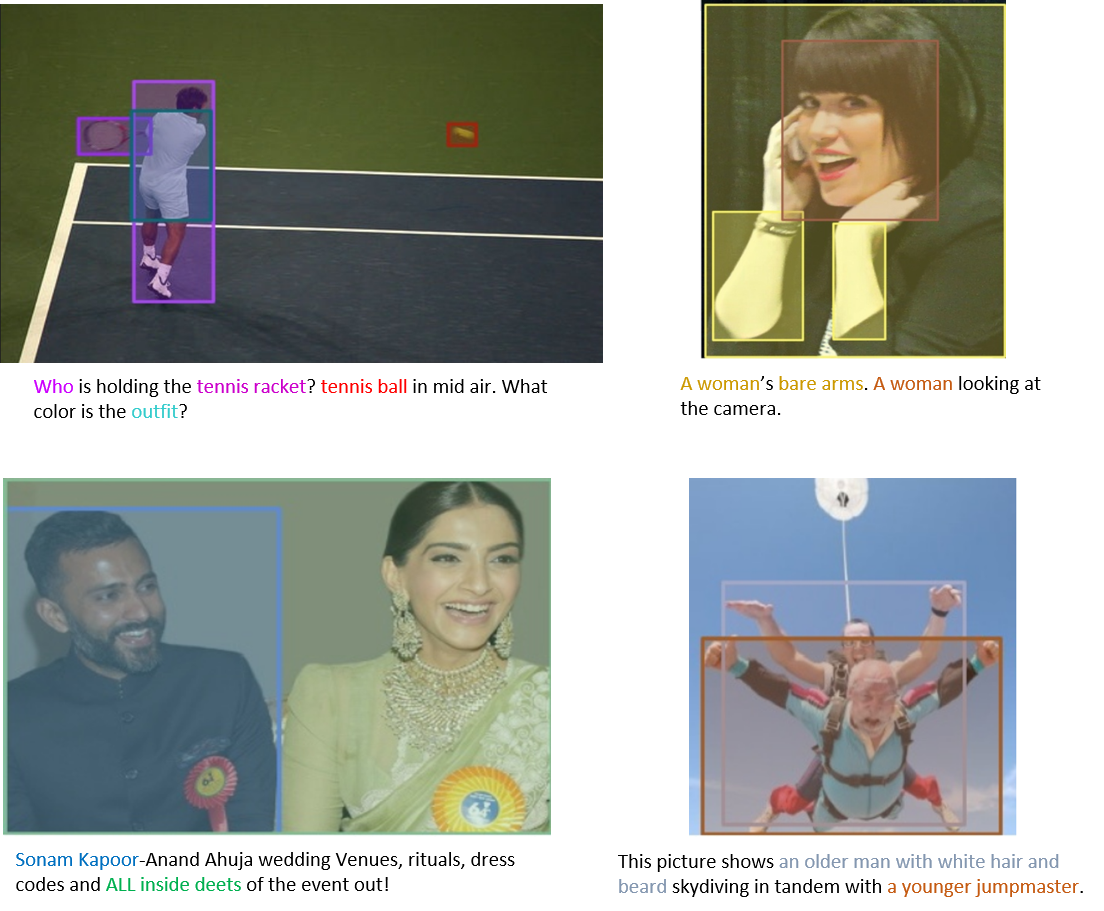}
    \end{center}
    \caption{Visualization of Pre-training Datasets. The first row displays images sourced from the GQA dataset, while the second row displays two images from the GRIT dataset. }
    \label{fig:vis_pre}
\end{figure}

\section{Visualization}

\subsection{Visualization on Pre-training Dataset}

In Figure \ref{fig:vis_pre}, we present visualizations of the pretraining datasets. Our analysis of these datasets revealed several noisy elements that could potentially undermine the training effectiveness. For example, some captions contain function words without substantive content, such as 'Who' in the top-left image and proper nouns like people's names in the bottom-left image. Additionally, the GRIT dataset, which utilizes GLIP for generating pseudo labels, may have inaccuracies in annotations. This is evident in the bottom-right image, where the box annotations appear to be incorrectly assigned. There are similar situations in GQA dataset. In the top-right image, the same phrase 'a woman' in a caption is assigned to different boxes, which contradicts the phrase grounding settings.  

\begin{table*}[ht]
    \centering
    \caption{Results on Flickr30K}
    \label{tab:flick}
    \begin{tabular}{l|lllllllll}
    \toprule
        Model & Pre-Train Data & Val R@1 & Val R@5 & Val R@10 & Test R@1 & Test R@5 & Test R@10 \\ 
        \hline
        GLIP-T & O365,GoldG & 84.9 & 94.9 & 96.3 & 85.6 & 95.4 & 96.7 \\ 
        GLIP-T & O365,GoldG,CC3M,SBU & 85.3 & 95.5 & 96.9 & 86.0 & 95.9 & 97.2 \\
        G-DINO-T(c) & O365,GoldG,Cap4M & 87.8 & 96.6 & 98.0 & 88.1 & 96.9 & 98.2 \\
        MM-G-T(b) & O365,GoldG & 85.5 & 95.6 & 97.2 & 86.2 & 95.7 & 97.4 \\
        MM-G-T(c1) & O365,GoldG,GRIT & 86.7 & 95.8 & 97.6 & 87.0 & 96.2 & 97.7\\
        MM-G-T(c2) & O365,GoldG,GRIT & 86.7 & 95.8 & 97.6 & 87.0 & 96.2 & 97.7\\
        MM-G-T(c3) & O365,GoldG,GRIT,V3Det & 86.7 & 96.0 & 97.6 & 87.2 & 96.2 & 97.7 \\
    \bottomrule
    \end{tabular}
\end{table*}

\begin{table*}[ht]
    \centering
    \caption{Zero-shot Results on ODinW35}
    \label{tab:ODinW35}
    \resizebox{\textwidth}{!}{%
    \begin{tabular}{l|ccccccc}
    \toprule
        Method &OdinW13 & ODinW35& G-DINO-T(c) & MM-G-T(b) & MM-G-T(c1)& MM-G-T(c2) & MM-G-T(c3) \\ 
        \hline
        AerialMaritimeDrone\_large& \Checkmark & \Checkmark & 0.173 & 0.133 & 0.155 & 0.177 & 0.151 \\
        AerialMaritimeDrone\_tiled & & \Checkmark& 0.206 & 0.170 & 0.225 & 0.184 &0.206 \\
        AmericanSignLanguageLetters & & \Checkmark& 0.002 & 0.016 & 0.020 & 0.011 & 0.007 \\
        Aquarium& \Checkmark & \Checkmark & 0.195 & 0.252 & 0.261 & 0.266 & 0.283 \\
        BCCD & & \Checkmark& 0.161 & 0.069 & 0.118 & 0.083 &0.077 \\
        boggleBoards & & \Checkmark& 0.000 & 0.002 & 0.001 &0.001 & 0.002 \\ 
        brackishUnderwater & & \Checkmark& 0.021 & 0.033 & 0.021 &0.025& 0.025 \\
        ChessPieces & & \Checkmark& 0.000 & 0.000 & 0.000 &0.000 & 0.000 \\
        CottontailRabbits& \Checkmark & \Checkmark & 0.806 & 0.771 & 0.810 &0.778 & 0.786 \\
        dice & & \Checkmark& 0.004 & 0.002 & 0.005 &0.001& 0.001 \\
        DroneControl & & \Checkmark& 0.042 & 0.047 & 0.097 &0.088 & 0.074 \\
        EgoHands\_generic & & \Checkmark& 0.608 & 0.527 & 0.537 &0.506& 0.519 \\
        EgoHands\_specific & & \Checkmark& 0.002 & 0.001 & 0.005 &0.007& 0.003 \\
        EgoHands& \Checkmark & & 0.608 & 0.499 & 0.537 & 0.506 & 0.519 \\
        HardHatWorkers & & \Checkmark& 0.046 & 0.048 & 0.070 &0.070& 0.108 \\
        MaskWearing & & \Checkmark& 0.004 & 0.009 & 0.004 &0.011& 0.009 \\
        MountainDewCommercial & & \Checkmark& 0.430 & 0.453 & 0.465 & 0.194& 0.430 \\
        NorthAmericaMushrooms& \Checkmark & \Checkmark & 0.471 & 0.331 & 0.462 &0.669& 0.767 \\
        openPoetryVision & & \Checkmark& 0.000 & 0.001 & 0.000 &0.000 & 0.000 \\
        OxfordPets\_by\_breed & & \Checkmark& 0.003 & 0.002 & 0.004 &0.006& 0.004 \\
        OxfordPets\_by\_species & & \Checkmark& 0.011 & 0.019 & 0.016 & 0.020&0.015 \\
        PKLot & & \Checkmark& 0.001 & 0.004 & 0.002 &0.008 & 0.007 \\
        Packages& \Checkmark & \Checkmark & 0.695 & 0.707 & 0.687 &0.710 &0.706 \\ 
        PascalVOC& \Checkmark & \Checkmark & 0.563 & 0.565 & 0.580 &0.566& 0.566 \\
        pistols& \Checkmark & \Checkmark & 0.726 & 0.585 & 0.709 &0.671 &0.729 \\
        plantdoc & & \Checkmark& 0.005 & 0.005 & 0.007 &0.008& 0.011 \\
        pothole& \Checkmark & \Checkmark & 0.215 & 0.136 & 0.219 &0.077& 0.168 \\
        Raccoons& \Checkmark & \Checkmark & 0.549 & 0.469 & 0.511 &0.553& 0.535 \\
        selfdrivingCar & & \Checkmark& 0.089 & 0.091 & 0.076 &0.094& 0.083 \\
        ShellfishOpenImages& \Checkmark & \Checkmark & 0.393 & 0.321 & 0.437 &0.519& 0.488 \\
        ThermalCheetah & & \Checkmark& 0.087 & 0.063 & 0.081 &0.030& 0.045 \\
        thermalDogsAndPeople& \Checkmark & \Checkmark & 0.657 & 0.556 & 0.603 &0.493& 0.543 \\
        UnoCards & & \Checkmark& 0.006 & 0.012 & 0.010 &0.009& 0.005 \\
        VehiclesOpenImages& \Checkmark & \Checkmark & 0.613 & 0.566 & 0.603 &0.614& 0.615 \\
        WildfireSmoke & & \Checkmark& 0.134 & 0.106 & 0.154 &0.042& 0.127 \\
        websiteScreenshots & & \Checkmark& 0.012 & 0.02 & 0.016 &0.016& 0.016 \\ \hline
        ODinW13 Average & & & \textbf{0.514} & \textbf{0.453} & \textbf{0.511} & \textbf{0.516} & \textbf{0.533} \\
        ODinW35 Average & & & \textbf{0.227} & \textbf{0.202} & \textbf{0.228} & \textbf{0.214} & \textbf{0.284} \\
    \bottomrule
    \end{tabular}
    }
\end{table*}

\begin{table*}[ht]
    \centering
    \caption{Open vocabulary fine-tune results on COCO benchmark. }
    \label{tab:ovd_finetune_coco}
    \begin{tabular}{lcc|ccc|ccc}
    \toprule
        Model & Backbone & Setting & \multicolumn{3}{c}{mAP} & \multicolumn{3}{c}{AP@50} \\ 
         & & & box & Base & Novel & box & Base & Novel \\ 
        \midrule
        MM-G-T(c3) & Swin-T & zero-shot & 51.1 & 48.4 & 58.9 & 66.7 & 64.0 & 74.2 \\ 
        MM-G-T(c3) & Swin-T & open vocabulary 1x & \textbf{57.2(+8.8)} & \textbf{56.1(+7.7)} & \textbf{60.4(+1.5)} & \textbf{73.6(+6.9)} & \textbf{73.0(+9.0)} & \textbf{75.3(+1.1)} \\
    \bottomrule
    \end{tabular}
\end{table*}

\begin{figure*}[ht]
  \centering
  \includegraphics[width=0.7\linewidth]{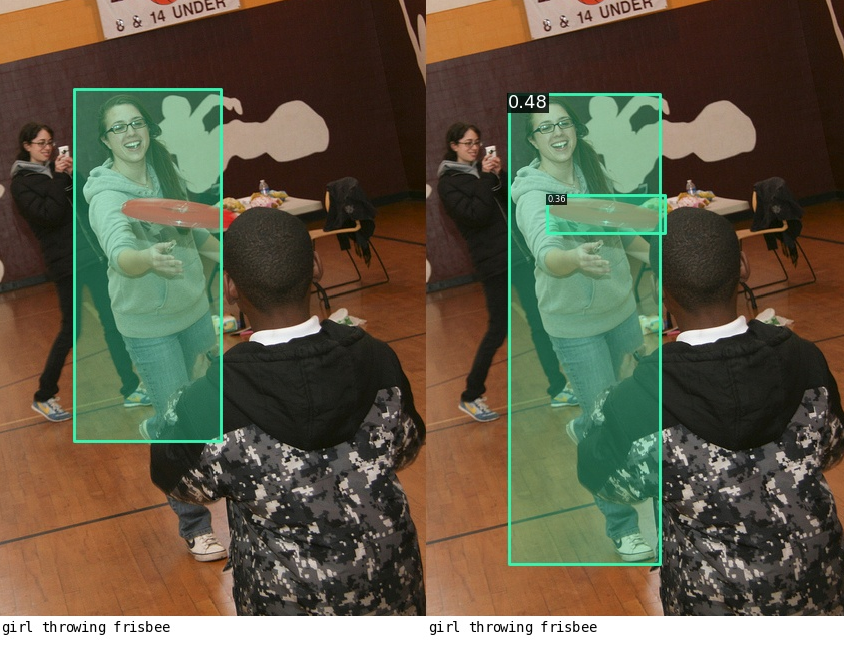}
  \caption{Comparison between ground-truth annotation and model's prediction. Concerning the 'girl' object, the prediction generated by MM-Grounding-DINO(right) appears to be more precise in contrast to the ground-truth annotations(left). \\}
  \label{fig:vis_lim_eva}
\end{figure*}

\subsection{Visualization on Model Predictions}
\begin{figure*}
\centering
\begin{subfigure}{\textwidth}
  \centering
  \includegraphics[width=\linewidth]{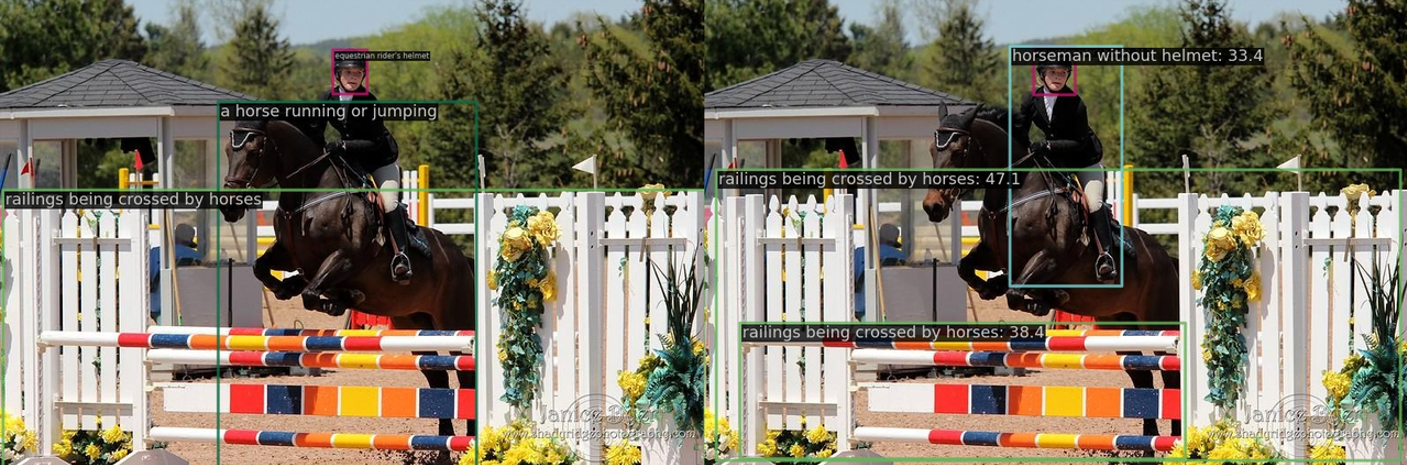}
  \caption{The prediction(right) primarily focuses on 'horseman' and 'helmet', while the key relational term ‘without’ was disregarded.}
  \label{fig:sub1}
\end{subfigure}
\newline

\begin{subfigure}{\textwidth}
  \centering
  \includegraphics[width=\linewidth]{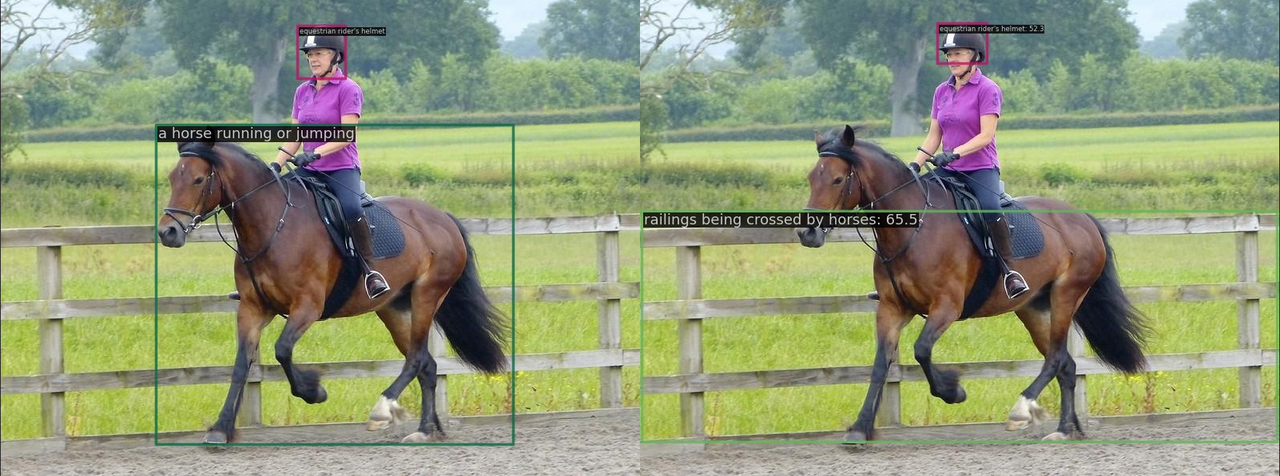}
  \caption{The prediction(right) incorrectly detected 'railings being crossed by horse'}
  \label{fig:sub2}
\end{subfigure}
\newline

\begin{subfigure}{\textwidth}
  \centering
  \includegraphics[width=\linewidth]{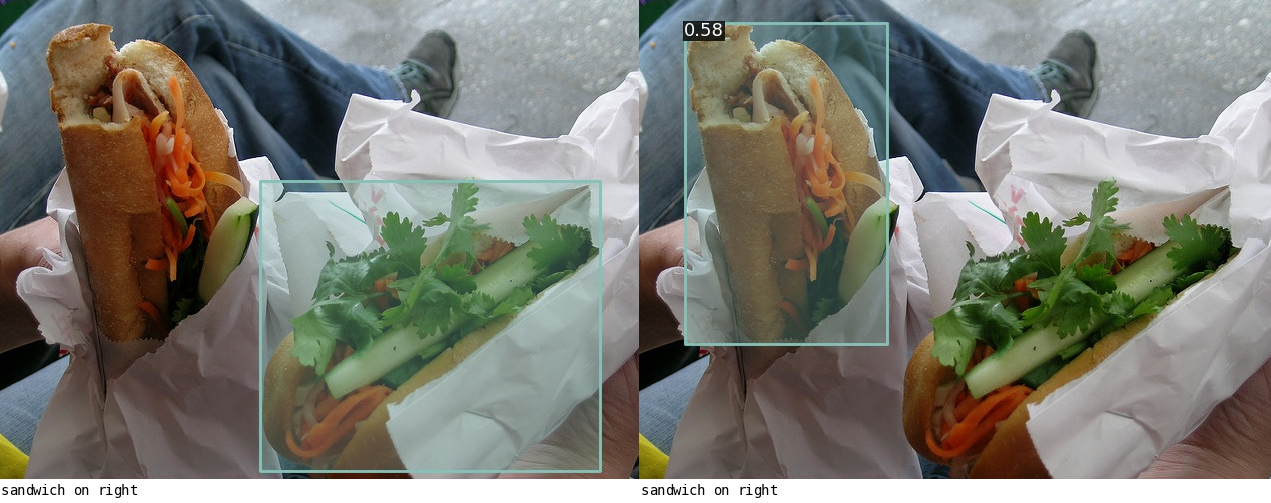}
  \caption{The prediction(right) confused the object on the left with the object on the right.}
  \label{fig:sub3}
\end{subfigure}
\newline



\caption{A series of five vertically aligned images.}
\label{fig:vis_premod}
\end{figure*}

For both Figure \ref{fig:vis_premod} and Figure \ref{fig:vis_lim_eva}, the ground-truth annotations are depicted on the left, while the predictions made by our model are presented on the right.\\

\noindent\textbf{Limitations of Evaluation. } Our visualization-based analysis of the evaluation process has revealed inaccuracies in ground-truth annotations of the evaluation dataset. This is evident in Figure \ref{fig:vis_lim_eva}, concerning the 'girl' object, our model's prediction appears to be more precise compared to the existing annotations. \\

\noindent\textbf{Limitations of Model. } During the pretraining phase, although the model has access to the entire caption, it tends to prioritize nouns, which are crucial for phrase grounding settings. For instance, in the caption 'horseman without helmet' depicted in Figure \ref{fig:sub1}, the model primarily focuses on 'horseman' and 'helmet', yet the key relational term ‘without’ was disregarded. This leads to an incapacity to differentiate between 'helmet' and 'without helmet'. Additionally, the model struggles with interpreting certain detailed descriptions, such as in Figure \ref{fig:sub2}, the model incorrectly detected 'railings being crossed by horse'. In terms of position description in the caption, the model only achieves suboptimal performance as shown in Figure \ref{fig:sub3}, which confused the object on the left with the object on the right. In Figure \ref{fig:vis_lim_eva}, our model additionally predicts 'frisbee' due to the phrase grounding settings, which leads to a lower performance in evaluation.

\end{document}